\documentclass[10pt,twocolumn,letterpaper]{article}

\usepackage{wacv}              

\usepackage{graphicx}
\usepackage{amsmath}
\usepackage{amssymb}
\usepackage{booktabs}
\usepackage{multirow}
\usepackage{xcolor}
\usepackage{amssymb}
\usepackage{pifont}

\definecolor{myRed}{rgb}{0.808,0.067,0.149}
\definecolor{myGreen}{rgb}{0.067,0.708,0.149}

\newcommand{\xmark}{{\color{myRed}\ding{55}}}%
\newcommand{\cmark}{{\color{myGreen}\ding{51}}}

\usepackage[pagebackref,breaklinks,colorlinks]{hyperref}

\usepackage[capitalize]{cleveref}
\crefname{section}{Sec.}{Secs.}
\Crefname{section}{Section}{Sections}
\Crefname{table}{Table}{Tables}
\crefname{table}{Tab.}{Tabs.}

\def\wacvPaperID{1708} 
\def\confName{WACV}
\def\confYear{2024}

\begin{document}

\title{\vspace*{-0.5cm}CL-MAE: Curriculum-Learned Masked Autoencoders\vspace*{-0.5cm}}

\author{Neelu Madan$^{1,\diamond}$, Nicolae-C\u{a}t\u{a}lin Ristea$^{2,3,\diamond}$, Kamal Nasrollahi$^{1,4}$,\\
Thomas B. Moeslund$^{1}$, Radu Tudor Ionescu$^{3,5,}$\thanks{corresp. author: raducu.ionescu@gmail.com; $^\diamond$equal contribution.}\\
$^1$Aalborg University, Denmark, $^2$University Politehnica of Bucharest, Romania,\\
$^3$University of Bucharest, Romania, $^4$Milestone Systems, Denmark, $^5$SecurifAI, Romania\vspace*{-0.3cm}
}

\maketitle
\begin{abstract}
\vspace*{-0.2cm}
Masked image modeling has been demonstrated as a powerful pretext task for generating robust representations that can be effectively generalized across multiple downstream tasks. Typically, this approach involves randomly masking patches (tokens) in input images, with the masking strategy remaining unchanged during training. In this paper, we propose a curriculum learning approach that updates the masking strategy to continually increase the complexity of the self-supervised reconstruction task. We conjecture that, by gradually increasing the task complexity, the model can learn more sophisticated and transferable representations. To facilitate this, we introduce a novel learnable masking module that possesses the capability to generate masks of different complexities, and integrate the proposed module into masked autoencoders (MAE). Our module is jointly trained with the MAE, while adjusting its behavior during training, transitioning from a partner to the MAE (optimizing the same reconstruction loss) to an adversary (optimizing the opposite loss), while passing through a neutral state.
The transition between these behaviors is smooth, being regulated by a factor that is multiplied with the reconstruction loss of the masking module. The resulting training procedure generates an easy-to-hard curriculum. We train our Curriculum-Learned Masked Autoencoder (CL-MAE) on ImageNet and show that it exhibits superior representation learning capabilities compared to MAE. The empirical results on five downstream tasks confirm our conjecture, demonstrating that curriculum learning can be successfully used to self-supervise masked autoencoders. We release our code at \url{https://github.com/ristea/cl-mae}.
\vspace*{-0.35cm}
\end{abstract}

\setlength{\abovedisplayskip}{3.0pt}
\setlength{\belowdisplayskip}{3.0pt}

\section{Introduction}
\label{introduction}

\begin{figure*}[t]
    \centering
    \includegraphics[width=0.94\textwidth]{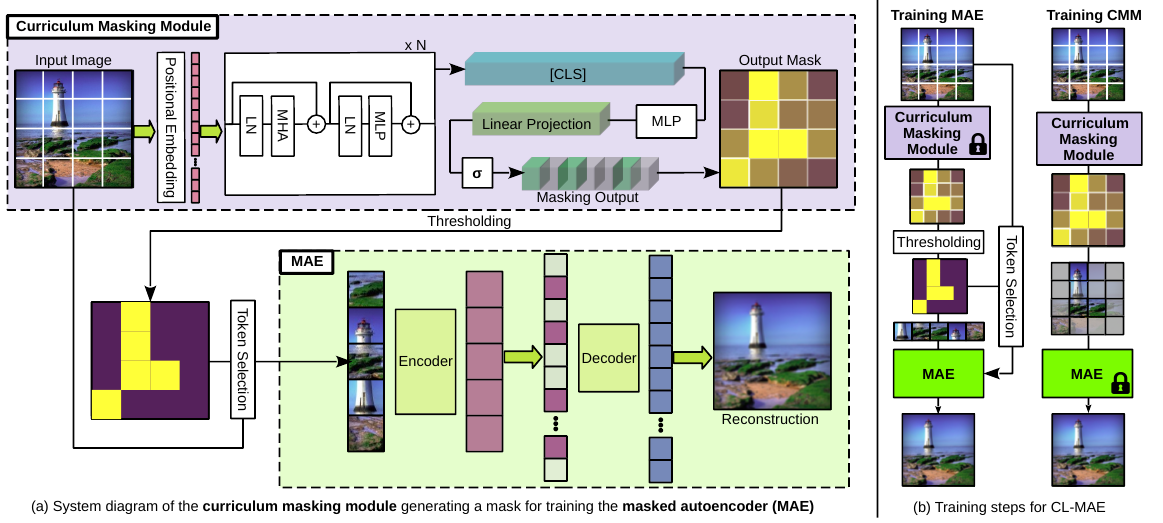}
    \vspace{-0.2cm}
    \caption{Our Curriculum-Learned Masked Autoencoder (CL-MAE) comprises a learnable masking module that decides what tokens need to be masked at each training iteration. The architecture of our module uses $N$ vision transformer (ViT) \cite{Dosovitskiy-ICLR-2021} blocks based on multi-head attention (MHA), layer normalization (LN) and multi-layer perceptrons (MLPs). The final [CLS] token is passed through an MLP, a linear projection and a sigmoid activation ($\sigma$), producing token masking probabilities. The masking module uses an easy-to-hard curriculum learning schedule that transitions smoothly from optimizing the same reconstruction objective as the MAE to an adversarial (opposed) objective. Hence, our masking module generates more or less complex masks, depending on its current objective. Our curriculum masking module (CMM) and the MAE \cite{He-CVPR-2022} are trained in alternating steps, similar to how generative adversarial networks \cite{Goodfellow-NIPS-2014} are trained. During inference, the masking module is removed. Best viewed in color.}\vspace{-0.2cm}
    \label{fig:system_diagram}
\end{figure*}

Self-supervised representation learning has grown to a prominent research topic, thanks to the possibility of learning representations that can be transferred to multiple visual tasks (referred to as downstream tasks), ranging from image recognition \cite{Dosovitskiy-ICLR-2021, Touvron-ICML-2021, Wu-ICCV-2021} and object detection \cite{Carion-ECCV-2020, Zheng-BMVC-2021, Zhu-ICLR-2020} to semantic segmentation \cite{Ghiasi-ECCV-2021, Freeseg-CVPR-2023, Wang-CVPR-2022, Das-WACV-2023}. These generic representations are usually learned by defining a self-supervised task, also known as \emph{pretext task}, where the labels are automatically generated from the available data, requiring no human supervision. Motivated by the achievements of masked language modeling techniques in natural language processing (NLP) \cite{Devlin-ACL-2019}, the field of computer vision has recently embraced masked image modeling as a self-supervised task \cite{Bao-ICLR-2022, Dong-AAAI-2023, He-CVPR-2016, Xie-CVPR-2022, Li-NeurIPS-2022, Wei-CVPR-2022, Zhang-NeurIPS-2022, Chen-CVPR-2023, Zhili-ICLR-2023, Shi-ICML-2022, Chen-ArXiv-2023}.
Masked image modeling involves masking a number of patches of an image and tasking the model at learning to reconstruct the masked information based on the remaining visible patches. Masked image models can be divided into two main categories with respect to reconstructing the target either as visual tokens \cite{Bao-ICLR-2022, Dong-AAAI-2023, Zhou-ICLR-2022, He-CVPR-2022, Zhang-NeurIPS-2022, Xie-CVPR-2022, Li-NeurIPS-2022} or features \cite{Wei-CVPR-2022, Wang-CVPR-2023}. The methods based on predicting masked tokens are the most prevalent ones, mostly because of their simplicity and better generalization capabilities. Even though a lot of attention has been dedicated to refining the pretext task \cite{Noroozi-CVPR-2018, Pathak-CVPR-2016, He-CVPR-2022, Zhang-ECCV-2016, Bao-ICLR-2022, Xie-CVPR-2022, Doersch-ICCV-2015, Balestriero-Meta-2023},  comparatively less attention has been paid to the token selection strategy \cite{Li-Arxiv-2022, Li-NeurIPS-2022, Chen-ArXiv-2023}. The mask selection criteria are often based on semantic object parts \cite{Li-NeurIPS-2022, Chen-ArXiv-2023} or uniform sampling \cite{Li-Arxiv-2022}. Unlike existing approaches, we propose to generate adaptive masks with different complexity levels, as part of the learning process, instead of using a single masking strategy. To this end, we propose a novel masking module, which is trained in an end-to-end fashion along with the MAE backbone \cite{He-CVPR-2022}. 
We also propose a novel curriculum learning setup, where the complexity of the pretext task is increased from easy-to-hard based on the generated masks, helping MAE to achieve better convergence and learn a more robust representation.  

We introduce curriculum learning as the core element of our proposed method, while using MAE \cite{He-CVPR-2022} as the underlying backbone for representational learning. Curriculum learning \cite{Bengio-ICML-2009} operates on the premise that models learn to solve tasks in the increasing order of their complexity, which helps to learn robust representations and enhance generalization capabilities. Different from existing curriculum methods \cite{Soviany-IJCV-2022}, we propose to create a curriculum by generating masks of increasing difficulty during training. To achieve this, we propose a novel masking module that is trained together with the MAE backbone, as shown in Figure \ref{fig:system_diagram}. In order to generate the progressive masks, from easy to hard, we introduce a curriculum loss function to train our new masking module, which shares the same objective as the pretext task. The complexity of the generated mask is governed by a factor that decides the weight of our curriculum loss function. First, the weight is set to a positive value in order to generate easy masks and facilitate learning the pretext task. The factor is decreased at every epoch, and even flips from positive values to negative values. When the curriculum loss weight reaches negative values, the masking module learns to increase the complexity of the pretext task by generating hard masks, acting as an adversary to the MAE backbone. Hence, our masking module starts with the same objective as the MAE, but gradually transforms into an adversary during training. This generates an easy-to-hard curriculum for the MAE. The architecture of our learnable curriculum masking module consists of a number of vision transformer (ViT) blocks \cite{Dosovitskiy-ICLR-2021}, as illustrated in Figure \ref{fig:system_diagram} (a). The masking probabilities are derived from the [CLS] token, after applying a multi-layer perceptron (MLP) and a sigmoid ($\sigma$) activation. A thresholding operation transforms the masking probabilities into binary values, which are subsequently used to select the tokens for the MAE. The thresholding operation is required to prevent having a trivial reconstruction task (multiplying the tokens with masking probabilities does not really hide the information, and the MAE can easily learn to rescale the pixel values to their original magnitude). Unfortunately, the thresholding operation also prevents gradient propagation. To overcome this limitation, in each training iteration, we alternate between training the MAE and the masking module, as shown in Figure \ref{fig:system_diagram} (b).


We conduct nearest neighbor, linear probing and few-shot linear probing experiments on five downstream image classification tasks, comparing the representation learning capabilities of MAE \cite{He-CVPR-2022} and CL-MAE, upon self-supervising both models on ImageNet \cite{Russakovsky-IJCV-2015}. The empirical results confirm the superior performance of our framework across the entire set of tasks and data sets. Moreover, we present ablation results to illustrate the utility of the various losses and components used by our novel masking module.


Our main contributions are summarized below:
\begin{itemize}
    \item \vspace{-0.15cm} We introduce curriculum learning into the MAE framework \cite{He-CVPR-2022} to learn robust representations.
    \item \vspace{-0.15cm} We propose a novel learnable masking module that is capable of generating adaptive masks, according to the desired complexity level.
    \item \vspace{-0.15cm} We present comprehensive results on five downstream tasks, showing that our curriculum-learned MAE outperforms MAE by significant margins. 
\end{itemize}

\section{Related Work}
\label{related_work}

\noindent
\textbf{Self-supervised representation learning}.
The most popular approaches among state-of-the-art self-supervised methods are based on contrastive learning \cite{He-CVPR-2020, Chen-ICML-2020, Grill-NeurIPS-2020, Pinheiro-NeurIPS-2020, Van-NeurIPS-2021} and masked image modeling \cite{Bao-ICLR-2022, Dong-AAAI-2023, He-CVPR-2016, Xie-CVPR-2022, Li-NeurIPS-2022, Wei-CVPR-2022, Zhang-NeurIPS-2022, Chen-CVPR-2023, Zhili-ICLR-2023, Zhou-ICLR-2022, Chen-ArXiv-2023}. Contrastive learning is based on pulling positive example pairs closer, while pushing negative pairs farther apart to learn robust representations. Since the number of negative pairs is usually very large, many approaches employ hard negative mining to parse negative image pairs. SimCLR \cite{Chen-ICML-2020} is based on end-to-end training and involves a simple one-to-one comparison with each negative instance. MoCo \cite{He-CVPR-2020} applies a different parsing technique, employing a momentum encoder to create a dynamic dictionary of negative samples. BYOL \cite{Grill-NeurIPS-2020} depends only on the positive pairs and eliminates the need for negative pairs. All these methods treat an image and its augmented versions as positive pairs, thus relying on heavy augmentation techniques. Masked image modeling represents a relatively simpler approach, where the masked regions of an image are reconstructed based on the visible image content. He \etal~\cite{He-CVPR-2022} showed that randomly masking a large number of image patches, \ie~$75\%$, results in a challenging pretext task, which generates a robust representation, eliminating the need for data augmentation.  

\noindent
\textbf{Masked image modeling.}
A sizeable amount of research nowadays in self-supervised representation learning is based on masked image modeling \cite{Bao-ICLR-2022, Dong-AAAI-2023, He-CVPR-2016, Xie-CVPR-2022, Li-NeurIPS-2022, Wei-CVPR-2022, Zhang-NeurIPS-2022, Chen-CVPR-2023, Zhili-ICLR-2023, Zhou-ICLR-2022, Chen-ArXiv-2023}, which is essentially inspired from masked language modeling, \eg~BERT \cite{Devlin-ACL-2019}. The mainstream methods based on masked image modeling can be categorized into approaches aiming to reconstruct visual tokens \cite{Bao-ICLR-2022, Dong-AAAI-2023, Zhou-ICLR-2022, He-CVPR-2022, Zhang-NeurIPS-2022, Xie-CVPR-2022, Li-NeurIPS-2022} or features \cite{Wei-CVPR-2022, Wang-CVPR-2023}. 

Preliminary studies focused on predicting visual tokens typically rely on an external tokenizer, which creates a visual codebook to reconstruct the target information. BeiT \cite{Bao-ICLR-2022} and PeCo \cite{Dong-AAAI-2023} are based on generating an offline visual codebook using variational autoencoders. Following DaLL-E \cite{Ramesh-ICML-2021a}, iBOT \cite{Zhou-ICLR-2022} later proposed an online tokenizer based on teacher networks generated via self-distillation. To mitigate the requirement of generating a visual codebook, Wei \etal~\cite{Wei-CVPR-2022} proposed to reconstruct Histogram-of-Oriented-Gradient (HOG) features for the masked region. These approaches are now replaced with more straightforward methods \cite{He-CVPR-2016, Xie-CVPR-2022, Zhang-NeurIPS-2022, Li-NeurIPS-2022, Chen-ArXiv-2023, Li-Arxiv-2022, Shi-ICML-2022} that try to directly reconstruct pixel values. He \etal~\cite{He-CVPR-2022} proposed an aggressive masking procedure, randomly hiding $75\%$ of the image patches, which seems to result in a very efficient and effective pretext task to learn robust and generic representations. Xie \etal~\cite{Xie-CVPR-2022} increased the complexity of the pretext task by increasing the patch size and reducing the decoder network to a single layer, as they claim that a harder pretext task leads to a better representation. 

A subcategory of methods based on using pixel-wise reconstruction \cite{Li-Arxiv-2022, Li-NeurIPS-2022, Chen-ArXiv-2023} has focused on different masking strategies to obtain robust representations. Li \etal~\cite{Li-Arxiv-2022} proposed a token selection strategy for the pyramid-based ViT, as the random selection of He \etal~\cite{He-CVPR-2022} does not seem to work in this case. Li \etal~\cite{Li-NeurIPS-2022} proposed semantically-guided masking, a framework that contains two modules, a self-supervised part generator, and a MAE \cite{He-CVPR-2022} for representation learning. Inspired by Li \etal~\cite{Li-NeurIPS-2022}, Chen \etal~\cite{Chen-ArXiv-2023} unified the part generator and MAE into a single differentiable framework. Different from these approaches, we propose a flexible masking strategy throughout the training process, where the masking depends on the desired complexity of the task, which varies from easy to hard in our case. We also introduce a novel masking module that can easily generate the masks for each level of complexity. Hence, our approach is based on a unique curriculum masking strategy, which is not encountered in existing methods. 

\noindent
\textbf{Curriculum learning.}
Curriculum learning, as introduced by Bengio \etal~\cite{Bengio-ICML-2009}, is a strategy aimed at organizing input data or tasks in a meaningful order, from easy to hard, to enhance the overall learning outcome. It consists of two main components: a curriculum criterion \cite{Bengio-ICML-2009} and a scheduling function \cite{Bengio-ICML-2009}. Approaches in curriculum learning can be categorized into easy-to-hard (standard curriculum) and hard-to-easy (anti-curriculum) paradigms \cite{Soviany-IJCV-2022}. In the easy-to-hard paradigm, tasks are presented to the model in increasing order of complexity \cite{Bengio-ICML-2009, Chen-ICCV-2015, Ionescu-CVPR-2016, Pentina-CVPR-2014, Shi-CSL-2015}, while the hard-to-easy paradigm reverses this order \cite{Shrivastava-CVPR-2016, Braun-EUSIPCO-2017}. Constructing a curriculum involves using either approaches based on external complexity measures, such as the degree of occlusion and the complexity of the shape \cite{Bengio-ICML-2009, Duan-ECCV-2020}, or self-paced learning techniques \cite{Jiang-ICML-2017, Weinshall-ICML-2018, Hacohen-ICML-2019, Kumar-NIPS-2010}, in which the neural network dynamically assesses the difficulty of training samples based on their loss. The scheduling function determines when and how to update the training process and can be categorized as discrete or continuous. Discrete schedulers \cite{Bengio-ICML-2009, Spitkovsky-NAACL-2010} sort and divide the data into discrete subsets, according to the curriculum criterion. Conversely, continuous schedulers \cite{Hacohen-ICML-2019, Platanios-NAACL-2019} provide a gradually increasing proportion of difficult training samples to the model. 
Our method incorporates an easy-to-hard continuous scheduler based on the complexity of the pretext task, where the reconstruction error of the model is used as a measure to construct the curriculum. Our novel curriculum learning strategy is deeply integrated within the proposed masking module. The module generates easy masks by hiding tokens with low reconstruction errors and hard masks by hiding tokens with high reconstruction errors.
The complexity of the task gradually increases during training from easy-to-hard, and, at some point, the masking module learns to produce extremely hard masks via adversarial training. 

\section{Method}
\label{Method}

We propose a curriculum learning approach together with a learnable masking module to train masked autoencoders \cite{He-CVPR-2022}. The proposed curriculum learning setup is aimed at achieving a robust representation that can be generalized over multiple data sets and visual tasks. The easy-to-hard curriculum is created through a novel masking module, which learns what tokens to mask in order to make the reconstruction task more or less complex. In the beginning of the training process, the masking module is trained with the same objective as the MAE to make the pretext task easier, \ie it learns to mask tokens that are easy to predict by the MAE. As the training progresses, we reduce the magnitude of the module's objective, essentially letting the module mask tokens at random. Further into the training process, we reverse the objective of the proposed masking module with respect to the MAE, which creates an adversarial learning environment where the MAE continues to learn to reconstruct masked tokens, while the masking module tries to hide the tokens that are difficult to reconstruct. The whole process described above is controlled via a curriculum loss factor $\lambda_{\mbox{\scriptsize{CL}}}$ that linearly decreases from a positive value to a negative value. This generates a smooth transition between behaviors of our learnable masking module, initially acting as a partner to the MAE, and gradually transitioning to a neutral state, and later, becoming an adversary to the MAE. Aside from the curriculum learning loss, we enforce the prediction of discriminative and diverse masks via additional loss components. In the following, we describe the proposed masking module and the loss functions used to train our module together with the MAE, in an end-to-end fashion. 


\subsection{Learnable Masking Module}

The core element of our framework is the masking module, a pivotal component in crafting masks of varying difficulty levels for the reconstruction task. Taking an input image $\boldsymbol{I} \in \mathbb{R}^{h \times w \times c}$, we divide it, as identically done for the standard MAE, into $n = (h\cdot w)/p^2$ non-overlapping patches, each patch having $p \times p$ pixels. Further, all patches are flattened and projected via a linear layer into the input tokens. To the existing tokens, we concatenate the learnable {[CLS]} token $\boldsymbol{C} \in \mathbb{R}^{d}$, obtaining the input tokens $\boldsymbol{T} \in \mathbb{R}^{(n + 1) \times d}$, where $d$ is the token dimension.

The input tokens are processed by the transformer-based module, which is inspired by the Vision Transformer (ViT) architecture \cite{Dosovitskiy-ICLR-2021}, obtaining the output tokens $\boldsymbol{T}_{out} \in \mathbb{R}^{(n + 1) \times d}$. Considering that the [CLS] token encapsulates information about the entire input image, we extract the output class token $\boldsymbol{C}_{out} \in \mathbb{R}^{d}$ and process it with a multi-layer perceptron (MLP) followed by a sigmoid activation function, as follows:
\begin{equation}
    \textbf{Z} = \sigma(\mbox{MLP}(\textbf{C}_{out})),
\end{equation}
where $\sigma$ denotes the sigmoid activation, and $\boldsymbol{Z} \in [0,1]^n$ is the soft output mask. Each element $z_i$ of $\boldsymbol{Z}$ represents the probability of keeping the $i$-th input token visible, \ie~inferring the corresponding token through the MAE for the current iteration. When training the masking module, the tensor $\boldsymbol{Z}$ is directly multiplied with the input tokens, allowing gradients to pass through our module. However, training the MAE with soft masking probabilities makes the reconstruction task much easier, \ie~the only job of MAE is to rescale the values of the softly masked tokens to counter the effect of multiplying the original tokens with the soft masking probabilities. Hence, before training the MAE, we apply a thresholding operation to transform the vector of soft masking probabilities into a binary masking vector $\boldsymbol{Z}^* \in \{0, 1\}^n$. Thanks to the Gaussian and Kullback-Leibler losses (detailed in Section \ref{sec_losses}), the threshold can be fixed to $0.5$ without tuning. During inference, the masking module is removed.

\subsection{Joint Training Procedure}

The output of the masking module is a soft vector with values between $0$ and $1$. This vector is transformed into a binary vector via a thresholding operation, which prevents gradient propagation (its gradients are equal to zero). Therefore, when we train the MAE backbone by masking input tokens, gradients with respect to the reconstruction loss are not propagated to our masking module. To alleviate this issue, we employ a two-step end-to-end training iteration, which independently updates the MAE and our masking module, similar to how generative adversarial networks are trained \cite{Goodfellow-NIPS-2014}. In the first step, we use the binary output $\boldsymbol{Z}^*$ of the frozen masking module to select visible tokens for the MAE backbone, leading to a conventional training step for the MAE. In the second step, we freeze the MAE backbone and train the masking module via $\boldsymbol{Z}$, replacing the thresholding and token selection operations with a multiplication operation. More precisely, instead of selecting which tokens should be passed to the MAE, we multiply all input tokens with the soft output of our masking block, allowing gradient propagation. In this fashion, we can propagate gradients with respect to the reconstruction loss of the MAE, transforming our module into a learnable component. We illustrate both training steps in Figure \ref{fig:system_diagram} (b).

\subsection{Proposed Loss Functions}
\label{sec_losses}


To train our learnable masking module, we propose four loss functions that are jointly minimized. We present the four losses and their roles below.


\noindent
\textbf{Curriculum loss.}
Taking inspiration from MAE \cite{He-CVPR-2022}, we employ the mean-squared error (MSE) metric between the normalized per-patch pixels of the reconstructed target ($\boldsymbol{\hat{I}}$) and the input image ($\boldsymbol{I}$) in our curriculum learning framework. The curriculum loss function is given by: 
\begin{equation}\label{eq:progressive_loss}
\begin{split}
    \mathcal{L}_{\mbox{\scriptsize{CL}}}(\boldsymbol{\hat{I}}, \boldsymbol{I}) &\!=\! 
 \lambda_{\mbox{\scriptsize{CL}}}^{(t)} \cdot (\boldsymbol{\hat{I}}\!-\! \boldsymbol{I})^2\!,
\end{split}
\end{equation}
where $\lambda_{\mbox{\scriptsize{CL}}}^{(t)} \in [-1,1]$ is linearly decreased at each training step $t \in \{0,1,2,...,T\}$, as follows: 
\begin{equation}
\begin{split}
    \lambda_{\mbox{\scriptsize{CL}}}^{(0)} &= 1,\\
    \lambda_{\mbox{\scriptsize{CL}}}^{(t+1)} &= \lambda_{\mbox{\scriptsize{CL}}}^{(t)} - k,
\end{split}
\end{equation}
where $k \in [0, 2/T]$ is a tunable decay value, and $T$ is the total number of training iterations. Note that $k$ determines how soon the masking module switches from a consensual objective to an adversarial one. When $k$ is set to the minimum value, there is no adversarial training. For the maximum decay $k=2/T$, the adversarial training starts halfway into the training process. Depending on $k$, the value of $\lambda_{\mbox{\scriptsize{CL}}}^{(T)}$ can be between $1$ and $-1$. As long as $\lambda_{\mbox{\scriptsize{CL}}}^{(t)} > 0$, the masking module tries to minimize $\mathcal{L}_{\mbox{\scriptsize{CL}}}$, contributing to simplifying the pretext task. When $\lambda_{\mbox{\scriptsize{CL}}}^{(t)}$ becomes negative, the masking module starts maximizing the difference between $\boldsymbol{\hat{I}}$ and $\boldsymbol{I}$, becoming an adversary to the MAE. By decreasing $\lambda_{\mbox{\scriptsize{CL}}}$ over time, our loss function constructs a curriculum that progresses from a simple to a challenging reconstruction task, thereby fostering a more effective representation learning process.

\noindent
\textbf{Gaussian loss.}
We incorporate a Gaussian objective into our masking module to enforce discriminative outputs. This objective forces the module to be decisive in picking which tokens should be or not be masked, pushing the masking probabilities away from $0.5$, towards $0$ or $1$. Thus, the soft output vector $\boldsymbol{Z}$ is expected to contain values close to $0$ for patches that need to be masked, and values close to $1$ for patches that must be kept visible. To achieve this behavior, we employ a Gaussian loss, as follows:
\begin{equation}
    \mathcal{L}_{\mbox{\scriptsize{Gauss}}} = \frac{1}{{\sigma \sqrt{2\pi}}} \exp\left(-\frac{{(\boldsymbol{Z} - \mu)^2}}{{2\sigma^2}}\right),
    \label{eq:gauss_loss}
\end{equation}
where $\mu$ denotes the mean, $\sigma$ corresponds to the standard deviation, and $\boldsymbol{Z}$ represents the output of the masking module. To push the masking probabilities away from $0.5$, we set $\mu$ to $0.5$. To regulate the resulting gradients, we use $\sigma=0.12$ in all our experiments. By training the masking module with the proposed Gaussian loss, it acquires the capability to determine whether a specific patch should be masked or left unmasked. This objective is meant to minimize the difference between the soft vector $\boldsymbol{Z}$ and the binary vector $\boldsymbol{Z}^*$. Hence, when we switch the thresholding operation applied to $\boldsymbol{Z}$ on and off to alternate between updating the weights of the MAE backbone and those of our module, a low difference between $\boldsymbol{Z}$ and $\boldsymbol{Z}^*$ makes the training steps more consistent with each other.

\noindent
\textbf{Kullback-Leibler loss.}
In our curriculum learning setup, the masking module can easily learn to shortcut the reconstruction task. For example, when the module aims to minimize the reconstruction error, it tends to avoid masking altogether. Conversely, when it behaves as an adversary and aims to maximize the reconstruction error, it tends to mask all the patches. To eliminate such shortcuts, we introduce a new loss function that aims to ensure a fixed number of tokens is always masked, regardless of the complexity of the reconstruction task. 
To enforce a predetermined masking ratio, we integrate a loss based on the Kullback-Leibler (KL) divergence into our learning framework. Our methodology involves generating distinct distributions for tokens that are masked and those that are visible, all based on a predefined masking ratio. This process includes creating two separate bins: one bin tallies the count of masked tokens, while the other bin keeps track of visible tokens. By establishing these distributions, our aim is to align with the target masking ratio. In order to gauge the difference between the intended distribution and the actual distribution of the outputs, we compute the KL divergence, which quantifies the dissimilarity between the two distributions. The loss based on the KL divergence is defined as follows:
\begin{equation}
\mathcal{L}_{\scriptsize{\mbox{KL}}} = m \cdot \log{\left(\frac{\hat{m}}{m}\right)} + v \cdot \log{\left(\frac{\hat{v}}{v}\right)},
\label{eq:kl_loss}
\end{equation}
where $\hat{m}$ represents the number of tokens to be masked estimated by the masking module, $m$ denotes the desired number of tokens to be masked (our target masking distribution), $\hat{v}$ signifies the estimated number of visible tokens, and $v$ the desired number of visible tokens. 
The loss is computed by evaluating the logarithmic ratios of the predicted outputs to the target values, weighted by the respective scale factors $m$ and $v$. The fractions $\frac{\hat{m}}{m}$ and $\frac{\hat{v}}{v}$ in Eq.~\eqref{eq:kl_loss} provide insights into the match between the predicted outputs and the target values. When the predicted outputs align with the target values, these fractions evaluate to $1$, and their logarithm to $0$. Hence, the loss value becomes $0$. The scale factors $m$ and $v$ ensure that each fraction is appropriately weighted.


\noindent
\textbf{Diversity loss.}
While the random masking process used by MAE \cite{He-CVPR-2022} inherently generates diverse masks, our learnable masking module might collapse to generating a single mask for all image samples. We thus need to employ a mechanism that ensures data diversity. To this end, we introduce a diversity loss that encourages the generation of different mask configurations. For a mini-batch of $p$ samples, the diversity loss is computed as follows:
\begin{equation}
    \mathcal{L}_{\scriptsize{\mbox{div}}} = \frac{1}{\frac{p \cdot (p-1)}{2}} \sum_{i=1}^{p} \sum_{j=i+1}^{p} {\exp} \left(- \lVert\boldsymbol{Z}_i - \boldsymbol{Z}_j\rVert^2\right),
    \label{eq:diversity_loss}
\end{equation}
\noindent
where $\boldsymbol{Z}_i$ and $\boldsymbol{Z}_j$ are the soft masking vectors corresponding to input images $\boldsymbol{I}_i$ and $\boldsymbol{I}_j$, respectively. Minimizing the proposed diversity loss is equivalent to maximizing the sum of distances between all pairs of soft masking vectors in a mini-batch. The loss is normalized with respect to the number of distinct image pairs in a mini-batch, obtaining a loss value that is independent of the batch size.

\noindent
\textbf{Joint Loss.}
The overall loss function used to optimize the masking module encompasses all the objectives presented so far, namely the curriculum loss ($\mathcal{L}_{\mbox{\scriptsize{CL}}}$), the Gaussian loss ($\mathcal{L}_{\mbox{\scriptsize{Gauss}}}$), the Kullback-Leibler loss ($\mathcal{L}_{\mbox{\scriptsize{KL}}}$), and the diversity loss ($\mathcal{L}_{\mbox{\scriptsize{div}}}$). Formally, the masking module is optimized via the following joint loss: 
\begin{equation}
    \mathcal{L}_{\mbox{\scriptsize{total}}} = \mathcal{L}_{\mbox{\scriptsize{CL}}} + \lambda_{\mbox{\scriptsize{Gauss}}} \!\cdot\!\mathcal{L}_{\mbox{\scriptsize{Gauss}}} + 
    \lambda_{\mbox{\scriptsize{KL}}} \!\cdot\!\mathcal{L}_{\mbox{\scriptsize{KL}}} + \lambda_{\mbox{\scriptsize{div}}} \!\cdot\!\mathcal{L}_{\mbox{\scriptsize{div}}}, 
    \label{eq:total_loss}
\end{equation}
where the hyperparameters $\lambda_{\mbox{\scriptsize{Gauss}}}>0$, $\lambda_{\mbox{\scriptsize{KL}}}>0$ and $\lambda_{\mbox{\scriptsize{div}}}>0$ dictate the contributions of the corresponding loss terms to the overall loss function. Note that the curriculum loss $\mathcal{L}_{\mbox{\scriptsize{CL}}}$ does not need a scaling factor, since it already includes one in its definition provided in Eq.~\eqref{eq:progressive_loss}.   

\section{Experiments and Results}

\subsection{Data Sets}

\noindent
\textbf{ImageNet.} The ImageNet bemchmark \cite{Russakovsky-IJCV-2015} contains over one million images from 1,000 categories, representing the most popular data set in computer vision. 

\noindent
\textbf{Aerial Images.} The Aerial Images data set (AID) \cite{Xia-TGRS-2017} comprises 10K aerial images of $600\times600$ pixels from 30 distinct categories, collected from Google Earth. The images are divided into 5K for training and 5K for testing.

\noindent
\textbf{Airbus Wind Turbines.} The Airbus Wind Turbines \cite{Airbus-Wind} data set comprises over 357K satellite images of $128\times128$ pixels, where the task is to classify images with and without wind turbines. We randomly split the data set into $80\%$ for training and $20\%$ for testing.

\noindent
\textbf{Architectural Heritage Elements.} The Architectural Heritage Elements (AHE) \cite{AHE_dataset} data set encompasses 10 distinct cultural heritage classes. This data set comprises 10,130 training images and 1,404 test images. The resolution of each image is $128\times128$ pixels.

\noindent
\textbf{Sea Animals.} The Sea Animals \cite{SeaAnimal_dataset} data set comprises images of 23 different sea creatures. This data set contains a total of 13,711 images of distinct resolutions. There are 12,339 images for training and 1,372 images for testing. 

\noindent
\textbf{Sport Balls.} The Sport Balls \cite{SportsBalls_dataset} data set is composed of 15 classes representing various sport balls. It incorporates 7,328 training images and 1,841 test images.

\subsection{Experimental Setup}

\noindent
\textbf{Backbones.} To compare the MAE and CL-MAE self-supervised training frameworks, we consider three ViT \cite{Dosovitskiy-ICLR-2021} backbones of different sizes, namely base (ViT-B), large (ViT-L) and huge (ViT-H). These backbones are already available in the official PyTorch repository\footnote{https://github.com/facebookresearch/mae} of MAE, which we employ in our experiments.

\noindent
\textbf{Evaluation protocols.} 
We start our experiments by self-supervising MAE \cite{He-CVPR-2022} and CL-MAE on 200 randomly chosen classes from ImageNet \cite{Russakovsky-IJCV-2015}. We then evaluate the learned representations on five downstream data sets, considering multiple evaluation scenarios: nearest neighbor, linear probing, and few-shot linear probing. In the first scenario, we apply a nearest neighbor model based on the Euclidean distance on top of the learned latent space. In the linear probing scenario, we train a Softmax layer on top of the learned encoders. The last scenario is similar to the second one, the only difference being the number of samples per class, which is restricted to a value in the set $\{1,2,4,8,16\}$. To better assess the power of the self-supervised representations, we refrain from fine-tuning the backbones on the downstream tasks. As evaluation metrics, we report the accuracy for the top-1 and top-5 predictions, denoted as Acc@1 and Acc@5, respectively. For the linear probing and few-shot linear probing protocols, we report the average accuracy rates over three runs for each model. This is not necessary for the nearest neighbors models, since they output deterministic predictions.

\begin{table}[t]
\centering 
\small{
\begin{tabular}{| l | c | c |  c |} 
\hline
{Method} & {$\lambda_{\mbox{\scriptsize{Gauss}}}$} & Acc@1 & Acc@5  \\
\hline
\hline
MAE (baseline) \cite{He-CVPR-2022} & - & $\mathbf{39.2}$ & $61.5$ \\
\hline
\multirow{4}{*}{CL-MAE} & $1$ & $35.0$ & $56.7$ \\
\multirow{4}{*}{(no curriculum)} & $2$ & $37.8$ & $59.6$ \\
                        & $5$ & $35.1$ & $56.7$ \\
                        & $10$ & ${38.7}$ & $\mathbf{61.7}$ \\
                        & $20$ & ${38.2}$ & ${61.3}$ \\
\hline
\end{tabular}
}
\vspace{-0.25cm}
\caption{ImageNet results while tuning the hyperparameter $\lambda_{\mbox{\scriptsize{Gauss}}}$ controlling the importance of our Gaussian loss. The results are obtained by nearest neighbor models applied on the self-supervised latent space of MAE and CL-MAE based on ViT-B. The curriculum loss is turned off. The top scores are in bold.}
\label{tab:ablation_gauss} 
\end{table}

\subsection{Hyperparameter Tuning}

For the vanilla MAE models, we use the hyperparameters recommended by He \etal \cite{He-CVPR-2022} for ImageNet. When we integrate our masking module, we do not change the recommended hyperparameters for the MAE backbones. However, there are some additional hyperparameters for our learnable masking module, which we tune on the ViT-B architecture. We reuse the hyperparameters established for our masking module on the ViT-L and ViT-H backbones.
    
\noindent
\textbf{Tuning for Gaussian loss.} For the moment, we turn off the curriculum learning, and focus on training the masking module to generate masking probabilities close to $0$ or $1$. Hence, we first tune the hyperparameter $\lambda_{\mbox{\scriptsize{Gauss}}}$, which controls the importance of the Gaussian loss. In Table \ref{tab:ablation_gauss}, we present preliminary results on ImageNet with various values for $\lambda_{\mbox{\scriptsize{Gauss}}}$ for the CL-MAE based on ViT-B. The empirical results indicate that the Gaussian loss is an important objective for our module, requiring a weight that is ten times greater than the other losses to produce optimal performance. We thus set $\lambda_{\mbox{\scriptsize{Gauss}}}=10$ in the subsequent experiments.

\begin{table}[t]
\centering 
\small{
\begin{tabular}{| l | c | c | c |} 
\hline
{Method} & {$\lambda_{\mbox{\scriptsize{KL}}}$} & Acc@1 & Acc@5 \\
\hline
\hline
MAE (baseline) \cite{He-CVPR-2022} & - & $39.2$ & $61.5$ \\
\hline                        
\multirow{4}{*}{CL-MAE}  & $0.1$ & $36.5$ & $59.3$ \\
\multirow{4}{*}{(no curriculum)} & $0.2$ & $37.8$ & $59.9$ \\
                                & $0.5$ & $38.2$ & $60.6$  \\
                                & $1$ & $\mathbf{39.5}$ & $\mathbf{61.9}$  \\
                                & $2$ & ${39.3}$ & ${61.7}$  \\
\hline
\end{tabular}
}
\vspace{-0.25cm}
\caption{ImageNet results while tuning the hyperparameter $\lambda_{\mbox{\scriptsize{KL}}}$ controlling the importance of our Kullback-Leibler loss. The results are obtained by nearest neighbor models applied on the self-supervised latent space of MAE and CL-MAE based on ViT-B. The curriculum loss is turned off. The top scores are in bold.}
\label{tab:ablation_kl} 
\end{table}

\noindent
\textbf{Tuning for Kullback-Leibler loss.} Next, we need to make sure that our module masks the right amount of patches, not more nor less. We keep the curriculum loss switched off, and tune the hyperparameter $\lambda_{\mbox{\scriptsize{KL}}}$, which represents the weight for the Kullback-Leibler loss. In Table \ref{tab:ablation_kl}, we present results on ImageNet with various values for $\lambda_{\mbox{\scriptsize{KL}}}$ for the CL-MAE based on ViT-B. Values below $1$ for $\lambda_{\mbox{\scriptsize{KL}}}$ produce considerable performance drops with respect to the MAE baseline. The reported accuracy rates show that the optimal weight for the Kullback-Leibler loss is $\lambda_{\mbox{\scriptsize{KL}}}=1$. We choose this setting ($\lambda_{\mbox{\scriptsize{KL}}}=1$) for the following experiments.

\begin{table}[t]
\centering 
\small{
\begin{tabular}{| l | c | c | c|} 
\hline
{Method} & {$\lambda_{\mbox{\scriptsize{CL}}}^{(T)}$} & Acc@1 & Acc@5 \\[1.5pt]
\hline
\hline
MAE (baseline) \cite{He-CVPR-2022} & - & $39.2$ & $61.5$ \\
\hline
\multirow{4}{*}{CL-MAE} & $0$ & $39.5$ & $61.9$ \\
 & $-0.1$ & $\mathbf{41.4}$ & $\mathbf{64.5}$ \\
 & $-0.15$ & $40.7$ & $63.6$ \\
 & $-0.2$ & $38.1$ & $60.4$ \\
\hline
\end{tabular}
}
\vspace{-0.25cm}
\caption{ImageNet results while tuning the hyperparameter $\lambda_{\mbox{\scriptsize{CL}}}^{(T)}$ (equivalent to tuning $k$) of our curriculum loss. The results are obtained by nearest neighbor models applied on the self-supervised latent space of MAE and CL-MAE based on ViT-B. The top scores are in bold.}
\vspace{-0.2cm}
\label{table:curriculum_interval} 
\end{table}

\noindent
\textbf{Tuning for curriculum loss.}
For the curriculum loss, we tune the decay value $k$ with respect to the last weight factor $\lambda_{\mbox{\scriptsize{CL}}}^{(T)}$, which specifies the importance of the adversarial objective in the last training iteration $T$. Since the relation between $k$ and $\lambda_{\mbox{\scriptsize{CL}}}^{(T)}$ is bijective, we express the tuning in terms of the more intuitive hyperparameter, namely $\lambda_{\mbox{\scriptsize{CL}}}^{(T)}$. We consider values for $\lambda_{\mbox{\scriptsize{CL}}}^{(T)}$ between $0$ and $-0.2$ and report the corresponding results in Table \ref{table:curriculum_interval}. Turning off the adversarial training, \ie setting $\lambda_{\mbox{\scriptsize{CL}}}^{(T)}=0$, leads to results that are marginally better than the vanilla MAE. In contrast, using too much adversarial training ($\lambda_{\mbox{\scriptsize{CL}}}^{(T)}=-0.2$) seems to actually harm the model. The results show that $\lambda_{\mbox{\scriptsize{CL}}}^{(T)}=-0.1$ is the optimal value for the CL-MAE based on ViT-B applied on ImageNet. Therefore, we preserve this value for the subsequent experiments.


\begin{table}[t]
\centering 
\small{
\begin{tabular}{| l | c | c | c |} 
\hline
{Method} & {$\lambda_{\mbox{\scriptsize{div}}}$} & Acc@1 & Acc@5  \\
\hline
\hline
MAE (baseline) \cite{He-CVPR-2022} & - & $39.2$ & $61.5$ \\
\hline
\multirow{3}{*}{CL-MAE} & $1$ & $41.4$  & $64.5$ \\
& $2$ & $\mathbf{42.1}$ & $\mathbf{65.1}$ \\
& $5$ & $40.6$ & $63.3$ \\
\hline
\end{tabular}
}
\vspace{-0.25cm}
\caption{ImageNet results while tuning the hyperparameter $\lambda_{\mbox{\scriptsize{div}}}$ controlling the importance of our diversity loss. The results are obtained by nearest neighbor models applied on the self-supervised latent space of MAE and CL-MAE based on ViT-B. The top scores are in bold.}
\label{table:ablation_div} 
\end{table}

\noindent
\textbf{Tuning for diversity loss.} Another important aspect is the diversity of the generated masks. The results presented so far used the default setting for the weight of the diversity loss, namely $\lambda_{\mbox{\scriptsize{div}}}=1$. However, our module might benefit from a higher emphasis on the diversity of the generated masks. To this end, we consider higher values for $\lambda_{\mbox{\scriptsize{div}}}$ and present the corresponding results in Table \ref{table:ablation_div}. The reported results indicate that $\lambda_{\mbox{\scriptsize{div}}}=2$ is the optimal choice, suggesting that the diversity of the generated masks is indeed important. We set $\lambda_{\mbox{\scriptsize{div}}}=2$ in the following experiments.

\begin{table}[t]
\centering 
\small{
\begin{tabular}{| l | c | c | c |} 
\hline
{Method} & {$N$} & Acc@1 & Acc@5 \\
\hline
\hline
MAE (baseline) \cite{He-CVPR-2022} & - & $39.2$ & $61.5$ \\
\hline
\multirow{3}{*}{CL-MAE} & $4$ & $42.1$ & $65.0$ \\
& $5$ & $\mathbf{42.1}$ & $\mathbf{65.1}$  \\
& $6$ & $40.0$ & $63.3$  \\
\hline
\end{tabular}
}
\vspace{-0.25cm}
\caption{ImageNet results while tuning the number of transformer blocks $N$ inside our learnable masking module. The results are obtained by nearest neighbor models applied on the self-supervised latent space of MAE and CL-MAE based on ViT-B. The top scores are in bold.}
\label{table:ablation_N} 
\end{table}

\noindent
\textbf{Number of transformer blocks.} 
Our masking module comprises a configurable number of transformer blocks $N$. We tune this hyperparameter considering values in the set $\{4,5,6\}$ and report the corresponding results in Table \ref{table:ablation_N}. The experiments show that choosing $N=5$ provides the best accuracy rates. Thus, we use $N=5$ in the next experiments.

\begin{table}[t]
\centering 
\setlength\tabcolsep{2.1pt}
\small{
\begin{tabular}{| l | c | c | c | c| c | c|} 
\hline
\multirow{2}{*}{Method} & \multicolumn{2}{c|}{ViT-B} & \multicolumn{2}{c|}{ViT-L} & \multicolumn{2}{c|}{ViT-H} \\
\cline{2-7}
   & Acc@1 & Acc@5  & Acc@1 & Acc@5 & Acc@1 & Acc@5 \\
\hline
\hline
MAE \cite{He-CVPR-2022} & $39.2$ & $61.5$               & $44.8$ & $67.0$ & $44.1$ & $66.1$  \\
CL-MAE & $\mathbf{42.1}$ & $\mathbf{65.1}$           & $\mathbf{45.2}$ & $\mathbf{67.2}$ & $\mathbf{45.7}$ & $\mathbf{68.2}$ \\
\hline
\end{tabular}
}
\vspace{-0.25cm}
\caption{ImageNet results obtained by nearest neighbor models applied on the self-supervised latent space of MAE \cite{He-CVPR-2022} and CL-MAE (ours) based on various backbones (ViT-B, ViT-L, ViT-H). The top scores for each backbone are in bold.}
\vspace{-0.2cm}
\label{tab:imgnet_main_results} 
\end{table}

\begin{table*}[t]
\centering 
\setlength\tabcolsep{4.9pt}
\small
\begin{tabular}{|c| l | c | c | c | c | c | c | c | c | c | c |} 
\hline
\multirow{3}{*}{Protocol} & \multirow{3}{*}{Method}  & \multicolumn{2}{c|}{\multirow{2}{*}{Aerial Images}}  & \multicolumn{2}{c|}{Airbus Wind}  & \multicolumn{2}{c|}{Architectural}  & \multicolumn{2}{c|}{\multirow{2}{*}{Sea Animals}} & \multicolumn{2}{c|}{\multirow{2}{*}{Sport Balls}} \\
&                         & \multicolumn{2}{c|}{} & \multicolumn{2}{c|}{{Turbines}}  & \multicolumn{2}{c|}{{Heritage Elements}}  & \multicolumn{2}{c|}{} & \multicolumn{2}{c|}{} \\
\cline{3-12}
&    & Acc@1 & Acc@5 & Acc@1 & Acc@5  & Acc@1 & Acc@5 & Acc@1 & Acc@5  & Acc@1 & Acc@5 \\ 
\hline
\hline
& MAE (Vit-B) \cite{He-CVPR-2022} & $80.2$ & $94.9$ & $92.1$ & $97.7$ & $76.8$ & $93.4$  & $51.2$ & $76.2$ & $57.6$ & $81.8$ \\
& CL-MAE (Vit-B) & $\mathbf{82.6}$ & $\mathbf{98.1}$ & $\mathbf{93.4}$ & $\mathbf{97.9}$ &  $\mathbf{79.3}$ & $\mathbf{94.0}$ & $\mathbf{56.2}$ & $\mathbf{79.7}$ & $\mathbf{60.2}$ & $\mathbf{84.5}$ \\
\cline{2-12}
Nearest & MAE (Vit-L) \cite{He-CVPR-2022} & $83.4$ & $95.7$ & $93.7$ & $98.5$ & $75.6$ & $93.1$ & $52.1$ & $77.0$ & $58.0$ & $82.5$ \\
Neighbor & CL-MAE (Vit-L) & $\mathbf{84.7}$ & $\mathbf{96.3}$ & $\mathbf{94.9}$ & $\mathbf{98.8}$ & $\mathbf{76.1}$ & $\mathbf{93.7}$ & $\mathbf{53.4}$ & $\mathbf{77.3}$ & $\mathbf{58.3}$ & $\mathbf{82.7}$ \\
\cline{2-12}
& MAE (Vit-H)  \cite{He-CVPR-2022} & $84.0$ & $97.9$ & $94.6$ & $98.4$ & $73.6$ & $92.2$ & $51.1$	& $76.8$ & $\mathbf{57.7}$ & $81.6$ \\
& CL-MAE (Vit-H) & $\mathbf{85.3}$ & $\mathbf{98.9}$ & $\mathbf{95.1}$ & $\mathbf{99.3}$ & $\mathbf{76.6}$ & $\mathbf{93.0}$ & $\mathbf{51.7}$ & $\mathbf{76.9}$ & $56.1$ & $\mathbf{82.0}$ \\
\hline
& MAE (Vit-B) \cite{He-CVPR-2022} & $84.8$ & $97.6$ & $97.8$ & $99.6$ & $84.6$ & $99.6$  & $67.5$ & $92.5$ & $62.9$ & $90.4$  \\
& CL-MAE (Vit-B) & $\mathbf{85.4}$ & $\mathbf{97.9}$ & $\mathbf{98.8}$ & $\mathbf{99.9}$ & $\mathbf{86.8}$ & $\mathbf{99.9}$  & $\mathbf{67.9}$ & $\mathbf{92.7}$ & $\mathbf{65.8}$ & $\mathbf{90.7}$  \\
\cline{2-12}
Linear & MAE (ViT-L) \cite{He-CVPR-2022} & $84.7$ & $97.7$ & $98.2$ & $99.6$ & $\mathbf{87.2}$ & $99.7$ & $69.8$ & $93.4$ & $67.4$ & $91.6$ \\
Probing & CL-MAE (ViT-L) & $\mathbf{85.9}$ & $\mathbf{98.1}$ & $\mathbf{99.1}$ & $\mathbf{99.9}$ & $\mathbf{87.2}$ & $\mathbf{99.8}$ & $\mathbf{71.0}$ & $\mathbf{93.8}$ & $\mathbf{69.0}$ & $\mathbf{92.1}$ \\
\cline{2-12}
& MAE (ViT-H) \cite{He-CVPR-2022} & $86.2$ & $98.5$ & $98.3$ & $99.7$ & $88.3$ & $\mathbf{99.7}$ & $75.4$ & $95.0$ & $74.6$ & $94.3$ \\
& CL-MAE (ViT-H) & $\mathbf{87.1}$ & $\mathbf{99.3}$ & $\mathbf{99.3}$ & $\mathbf{99.9}$ & $\mathbf{89.6}$ & $99.5$  & $\mathbf{76.2}$ & $\mathbf{95.2}$ & $\mathbf{75.1}$ & $\mathbf{94.9}$ \\
\hline
\end{tabular}
\vspace{-0.25cm}
\caption{Nearest neighbor (top half) and linear probing (bottom half) results on five benchmarks: Aerial Images, Airbus Wind Turbines, Architectural Heritage Elements, Sea Animals, and Sport Balls. The results are reported for MAE \cite{He-CVPR-2022} and CL-MAE (ours) based on various backbones (ViT-B, ViT-L, ViT-H). The top scores for each backbone on each data set are in bold.}
\vspace{-0.2cm}
\label{table:down_main_results} 
\end{table*}

\subsection{Results}

\noindent
\textbf{Results on ImageNet.} In Table \ref{tab:imgnet_main_results}, we present the nearest neighbor results for MAE \cite{He-CVPR-2022} and CL-MAE (ours) using three different backbones for each of the two frameworks, on the ImageNet data set. CL-MAE is consistently better than MAE \cite{He-CVPR-2022}, regardless of the underlying architecture. Our framework brings considerable performance gains for the ViT-B and ViT-H architectures, and moderate gains for ViT-L. According to a battery of paired McNemar's tests, all our gains are statistically significant, at a p-value of $0.001$.

\begin{table}[t]
\centering 
\setlength\tabcolsep{4.5pt}
\small
\begin{tabular}{|l| c | c |  c | c | c | c |} 
\hline
Method & $\mathcal{L}_{\mbox{\scriptsize{Gauss}}}$ & $\mathcal{L}_{\mbox{\scriptsize{KL}}}$ & $\mathcal{L}_{\mbox{\scriptsize{div}}}$ & $\mathcal{L}_{\mbox{\scriptsize{CL}}}$ &  Acc@1 & Acc@5  \\
\hline
\hline
MAE \cite{He-CVPR-2022} & \xmark & \xmark & \xmark & \xmark & $39.2$ & $61.5$ \\
\hline
\multirow{4}{*}{CL-MAE}& \cmark & \xmark & \xmark & \xmark & $38.7$ & $61.7$ \\
& \cmark & \cmark & \xmark & \xmark & $39.5$ & $61.9$ \\
& \cmark & \cmark & \cmark & \xmark & $40.8$ & $62.4$ \\
& \cmark & \cmark & \cmark & \cmark & $\mathbf{42.1}$ & $\mathbf{65.1}$ \\
\hline
\end{tabular}
\vspace{-0.25cm}
\caption{Ablation study on the loss functions used to train our learnable masking module. The results are obtained by nearest neighbor models applied on the self-supervised latent space of MAE and CL-MAE based on ViT-B. The top scores are in bold.}
\vspace{-0.2cm}
\label{tab:ablation_losses} 
\end{table}

\noindent
\textbf{Results on downstream tasks.} In Table \ref{table:down_main_results}, we present nearest neighbor and linear probing results on Aerial Images \cite{Xia-TGRS-2017}, Airbus Wind Turbines \cite{Airbus-Wind}, Architectural Heritage Elements \cite{AHE_dataset}, Sea Animals \cite{SeaAnimal_dataset}, and Sport Balls \cite{SportsBalls_dataset} data sets. The goal of these experiments is to assess the transferability of the self-supervised representations learned by MAE \cite{He-CVPR-2022} and CL-MAE, using three different backbones (ViT-B, ViT-L and ViT-H). In 57 out of 60 cases, CL-MAE outperforms MAE, with absolute gains varying between $+0.1\%$ and $+4.0\%$. A battery of paired McNemar's tests confirms that our gains are statistically significant, at a p-value of $0.001$. For the nearest neighbor protocol, there are 14 out of 30 cases where the absolute gains are higher than $1\%$. For the linear probing protocol, we have 7 out of 30 cases in which the gains are higher than $1\%$. The few-shot linear probing experiments (presented in the supplementary) are consistent with the results presented in Table \ref{table:down_main_results}. In summary, we conclude that our experiments provide comprehensive evidence indicating that CL-MAE is able to learn superior representations compared to MAE \cite{He-CVPR-2022}.

\noindent
\textbf{Ablation study.}
To assess the individual performance impact of the proposed loss functions, we conduct an ablation study on ImageNet and present the results in Table \ref{tab:ablation_losses}. Note that our learnable masking module needs at least one loss to function. Using solely the Gaussian loss is slightly worse than employing the vanilla MAE. Adding the Kullback-Leibler loss regulates the number of masked patches and improves the model, which becomes marginally better than MAE. The diversity loss plays an important role in further boosting the performance of CL-MAE. The curriculum loss also brings significant performance gains. In summary, the ablation study shows that each and every loss function contributes to the high performance gains of our model.

\section{Conclusion}
\label{conclusion}

In this paper, we introduced a novel approach for self-supervised representation learning with masked autoencoders, leveraging the concept of curriculum learning. Our method involves generating masks of increasing complexity using a novel learnable masking module. We proposed four losses to ensure that our masking module learns to produce masks that are decisive (close to binary), diverse, and in line with the imposed masking ratio and complexity. Our masking module is jointly trained with MAE, but its reconstruction objective changes during training, from a consensual objective (aiming to help MAE) to an adversarial objective (aiming to confuse MAE), generating an easy-to-hard curriculum learning setup. We conducted comprehensive experiments to compare our framework (CL-MAE) with the vanilla MAE. Our empirical results showed that CL-MAE learns better representations, outperforming the transfer learning capability of MAE. In future work, we aim to extend our analysis to other domains where MAE was successfully employed, \eg~video \cite{Tong-NeurIPS-2022} and audio-video \cite{Georgescu-ICCV-2023}.

\section{Acknowledgments}

This work has been funded by Milestone Systems through the Milestone Research Programme at AAU.

{\small
\bibliographystyle{ieee_fullname}
\bibliography{references}
}

\clearpage

\section{Supplementary}

In the supplementary, we analyze the behavior of our loss functions from a qualitative perspective, and present few-shot linear probing results on five downstream tasks. 

\begin{figure}[t]
    \centering
    \includegraphics[width=0.46\textwidth]{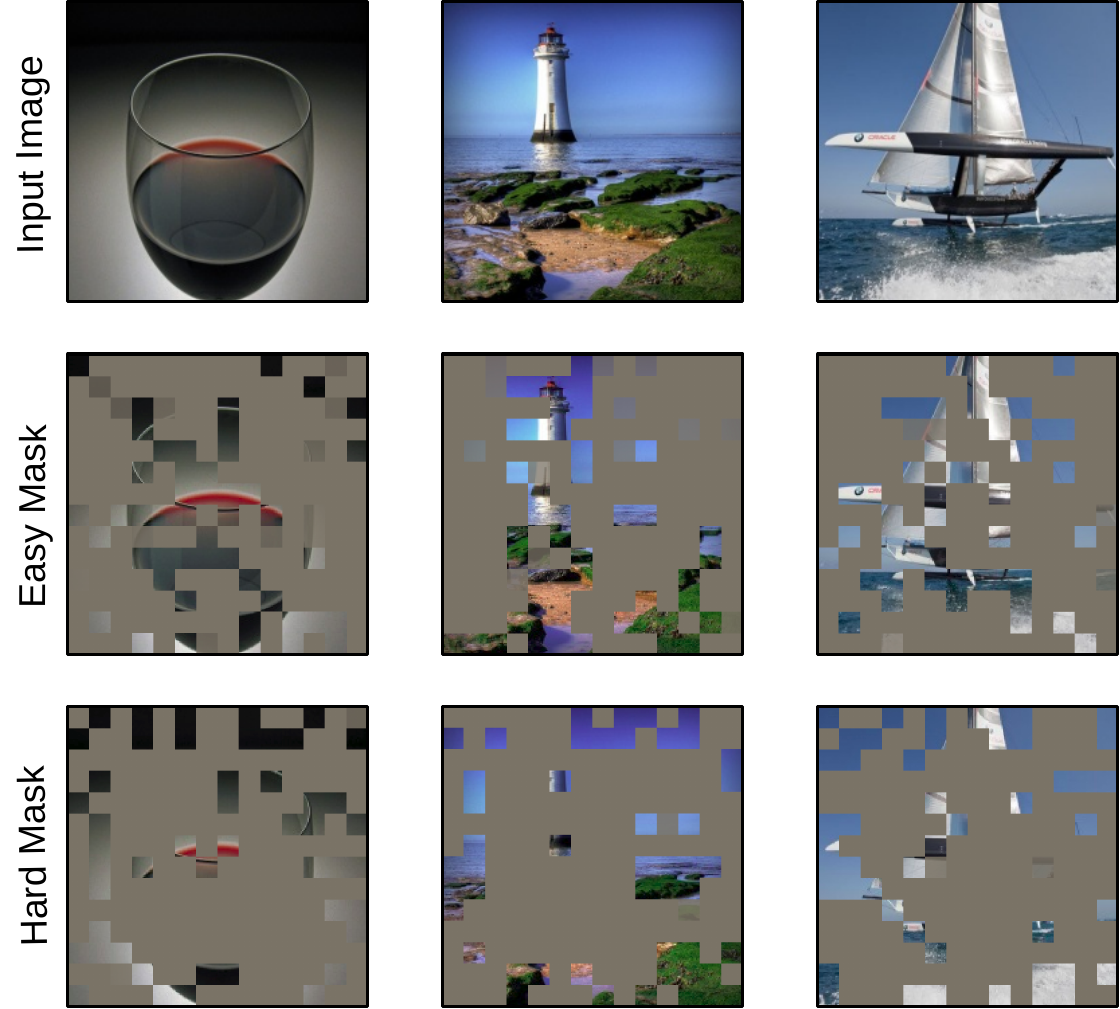}
    \caption{Masks generated by our masking module at two different moments during training, when all losses are in place, for the images on the top row. The masks on the second row are generated halfway during training, when the masking module is still acting as a partner to the MAE. In contrast, the masks on the bottom row are generated in the last epoch, when the masking module is behaving as an adversary to the MAE. Our module shifts its preference from masking non-salient tokens to masking tokens situated on edges and object contours, generating an easy-to-hard curriculum for the MAE.}
    \label{fig:curriculum_masks}
\end{figure}

\noindent
\textbf{Effect of curriculum learning on masking.} In Figure \ref{fig:curriculum_masks}, we provide some masks generated by our learnable masking module at two different states during the training process on ImageNet \cite{Russakovsky-IJCV-2015}. In the first state (second row), the masking module acts as a partner to the MAE, helping to ease the reconstruction task. In this state, the module seems to mask tokens from non-salient or plain texture regions, which can be easily inferred from surrounding patches. We observe that the object contours are mostly visible in the easy masks. In the second state (third row), our masking module acts as an adversary to the MAE, aiming to make the reconstruction task harder for the MAE. In this state, we observe that the module prefers to mask patches located on edges, object contours and salient regions. These patches are much harder to reconstruct based on the visible information. In summary, the examples presented in Figure \ref{fig:curriculum_masks} confirm that our learnable masking module behaves as expected, creating an easy-to-hard curriculum for the MAE.

\begin{figure}[t]
    \centering
    \includegraphics[width=0.48\textwidth]{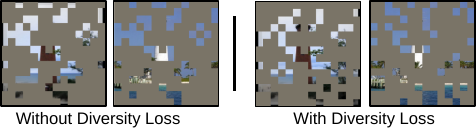}
    \caption{Masks generated by the proposed masking module without (left) and with (right) adding the diversity loss ($\mathcal{L}_{div}$). If the diversity loss ($\mathcal{L}_{div}$) is not included, the masking module can enter mode collapse and produce nearly identical masks. This can lead to overfitting CL-MAE on reconstructing certain patch configurations. The effect is no longer observed when the proposed diversity loss is employed.}
    \label{fig:diversity_loss}
\end{figure}

\begin{figure}[t]
    \centering
    \includegraphics[width=0.45\textwidth]{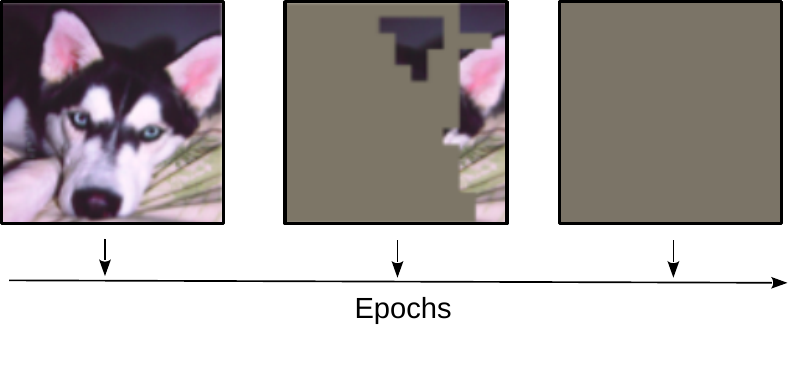}
    \vspace{-0.5cm}
    \caption{Masks generated by the proposed masking module during training, without the Kullback-Leibler loss. The masks evolve from leaving all patches visible (to reduce the reconstruction error for the MAE) to hiding all patches (to increase the reconstruction error for the MAE). The Kullback-Leibler loss is required to make sure the model always masks the desired number of patches.}
    \label{fig:kl_loss}
\end{figure}

\begin{figure*}[th]
    \centering  
    \begin{subfigure}{1\textwidth}
        \centering
    \includegraphics[width=1\linewidth]{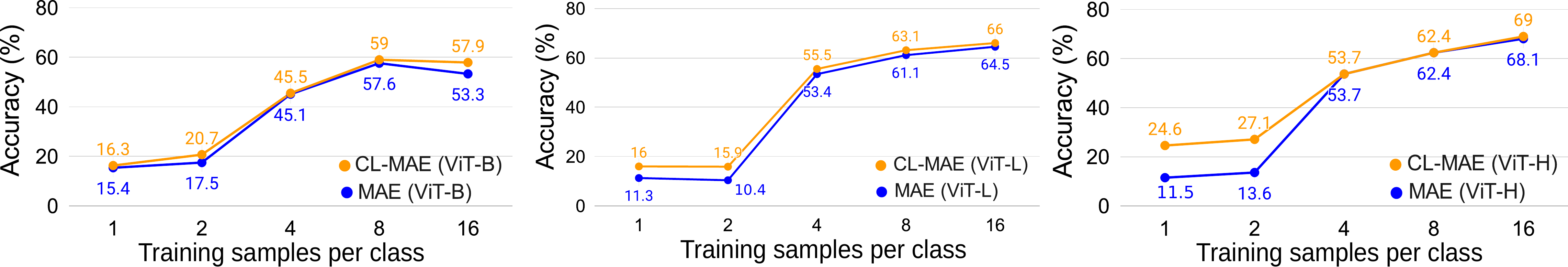}
        \caption{Few-shot linear probing results on Aerial Images \cite{Xia-TGRS-2017} with ViT-B (left), ViT-L (center) and ViT-H (right).}
        \label{fig:AID}
    \end{subfigure}
    \hfill
    \begin{subfigure}{1\textwidth}
        \centering
	    \includegraphics[width=\linewidth]{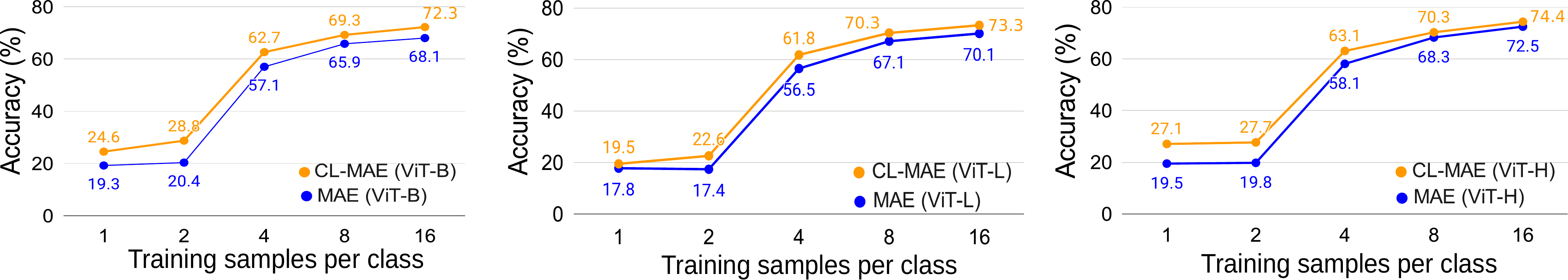}
        \caption{Few-shot linear probing results on Airbus Wind Turbines \cite{Airbus-Wind} with ViT-B (left), ViT-L (center) and ViT-H (right).}
        \label{fig:AWT}
    \end{subfigure}
    \begin{subfigure}{1\textwidth}
        \centering
	\includegraphics[width=\linewidth]{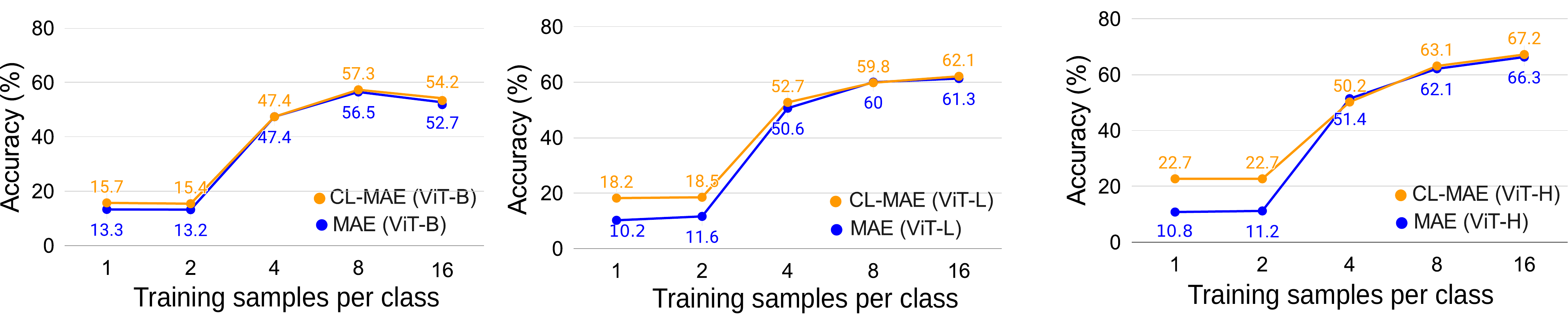}
        \caption{Few-shot linear probing results on Architectural Heritage Elements \cite{AHE_dataset} with ViT-B (left), ViT-L (center) and ViT-H (right).}
        \label{fig:AHE}
    \end{subfigure}
    \begin{subfigure}{1\textwidth}
        \centering
	\includegraphics[width=\linewidth]{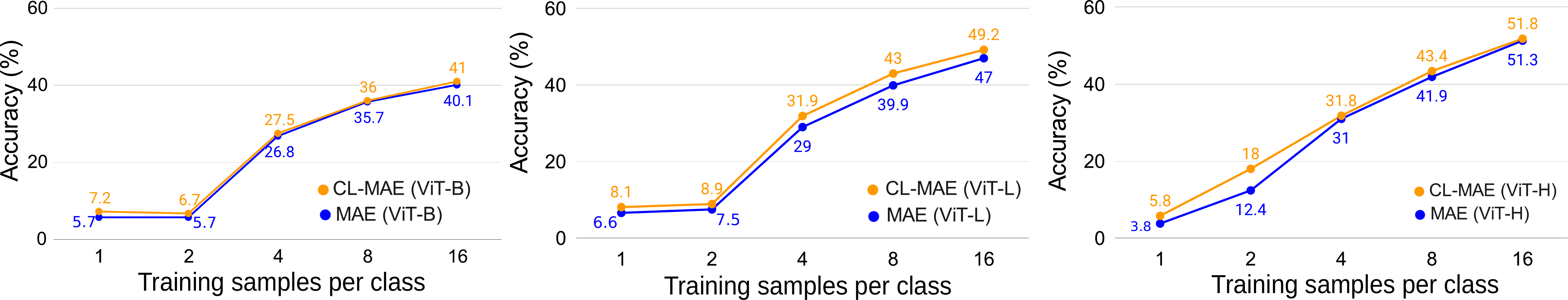}
        \caption{Few-shot linear probing results on Sea Animals \cite{SeaAnimal_dataset} with ViT-B (left), ViT-L (center) and ViT-H (right).}
        \label{fig:SeaAnimals}
    \end{subfigure}
    \begin{subfigure}{1\textwidth}
        \centering
	\includegraphics[width=\linewidth]{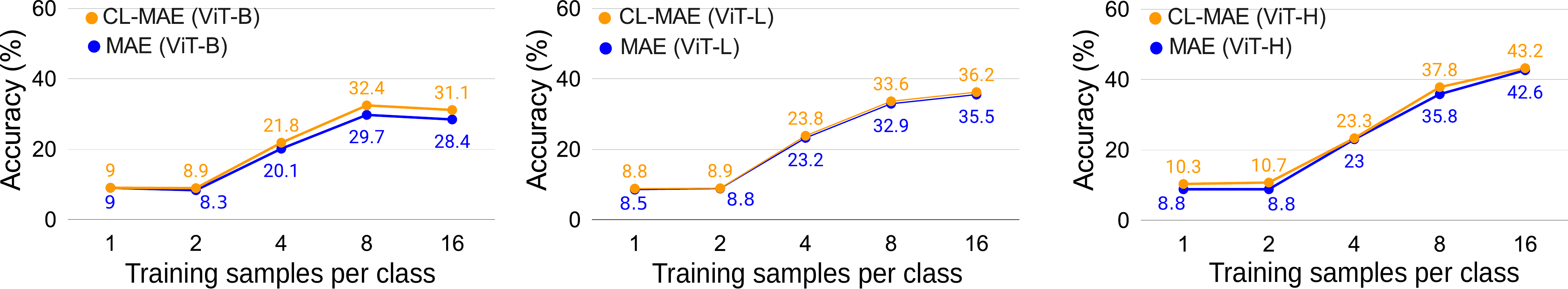}
        \caption{Few-shot linear probing results on Sport Balls \cite{SportsBalls_dataset} with ViT-B (left), ViT-L (center) and ViT-H (right).}
        \label{fig:SportsBall}
    \end{subfigure}
    \caption{Few-shot linear probing results for MAE \cite{He-CVPR-2022} and CL-MAE (ours) based on various backbones (ViT-B, ViT-L, ViT-H). The number of training samples per class varies between $1$ and $16$. The reported accuracy rates are averaged over three runs. Best viewed in color.}
    \label{fig:fewshot_results}
\end{figure*}

\begin{table*}[th]
\centering 
\setlength\tabcolsep{4.0pt}
\small
\begin{tabular}{|c| l | c | c | c | c | c | c | c | c | c | c |} 
\hline
\multirow{2}{*}{Few-shot}& \multirow{3}{*}{Method}  & \multicolumn{2}{c|}{\multirow{2}{*}{Aerial Images}}  & \multicolumn{2}{c|}{Airbus Wind}  & \multicolumn{2}{c|}{Architectural}  & \multicolumn{2}{c|}{\multirow{2}{*}{Sea Animals}} & \multicolumn{2}{c|}{\multirow{2}{*}{Sport Balls}} \\
\multirow{2}{*}{scenario} &                         & \multicolumn{2}{c|}{} & \multicolumn{2}{c|}{{Turbines}}  & \multicolumn{2}{c|}{{Heritage Elements}}  & \multicolumn{2}{c|}{} & \multicolumn{2}{c|}{} \\
\cline{3-12}
 &    & Acc@1 & Acc@5 & Acc@1 & Acc@5  & Acc@1 & Acc@5 & Acc@1 & Acc@5  & Acc@1 & Acc@5 \\ 
\hline
\hline
\multirow{6}{*}{1-shot} & MAE (Vit-B) \cite{He-CVPR-2022} & 15.4 & 48.6 & 19.3 & 54.3 & 13.3 & 47.2  & 5.7 & 20.6 & \textbf{9.0} & 32.6 \\
& CL-MAE (Vit-B) & \textbf{16.3} & \textbf{58.9} & \textbf{24.6} & \textbf{63.9} & \textbf{15.7} & \textbf{61.7} & \textbf{7.2} & \textbf{26.7}  & \textbf{9.0} & \textbf{34.5}  \\
\cline{2-12}
& MAE (Vit-L) \cite{He-CVPR-2022} & 11.3 & 39.8 & 17.8 & 56.2 & 10.2 & 40.5  & 6.6 & 27.7 & 8.5 & 34.9 \\
& CL-MAE (Vit-L) & \textbf{16.0} &  \textbf{50.7} & \textbf{19.5} & \textbf{63.8} & \textbf{18.2} & \textbf{55.8}  & \textbf{8.1} & \textbf{31.6} & \textbf{8.8} & \textbf{36.9} \\
\cline{2-12}
& MAE (Vit-H)  \cite{He-CVPR-2022} & 11.5 & 54.1 & 19.5 & 57.2 & 10.8 & 51.6  & 3.8 & \textbf{22.8} & 8.8 & 34.8 \\
& CL-MAE (Vit-H) &  \textbf{24.6} &  \textbf{58.2} & \textbf{27.1} & \textbf{59.0} & \textbf{22.7} & \textbf{56.0}  & \textbf{5.8} & 22.4 & \textbf{10.3} & \textbf{35.7} \\
\hline
\multirow{6}{*}{2-shot} & MAE (Vit-B) \cite{He-CVPR-2022} & 17.5 & 53.1 & 20.4 & 65.3 & 13.2 & 51.4 & 5.7 & 24.6  & 8.3 & 34.9 \\
& CL-MAE (Vit-B) &  \textbf{20.7} &  \textbf{60.3} & \textbf{28.8} & \textbf{69.9} & \textbf{15.4} & \textbf{60.2}  & \textbf{6.7}  & \textbf{25.8} & \textbf{8.9} & \textbf{37.2} \\
\cline{2-12}
& MAE (Vit-L) \cite{He-CVPR-2022} & 10.4 & 39.6 & 17.4 & 60.1 & 11.6 & 42.9  & 7.5 & \textbf{33.6} & 8.8 & 32.9 \\
& CL-MAE (Vit-L) &  \textbf{15.9} &  \textbf{51.3} & \textbf{22.6} & \textbf{66.2} & \textbf{18.5} & \textbf{53.5}  & \textbf{8.9} & 31.9 & \textbf{8.9} & \textbf{34.7} \\
\cline{2-12}
& MAE (Vit-H)  \cite{He-CVPR-2022} & 13.6 & 50.7 & 19.8 & 61.7 & 11.2 & 51.5  & 12.4 & 43.0 & 8.8 & 34.9 \\
& CL-MAE (Vit-H) &  \textbf{27.1} &  \textbf{58.8} & \textbf{27.7} & \textbf{69.2} & \textbf{22.7} & \textbf{55.8}  & \textbf{18.0} & \textbf{52.2} & \textbf{10.7} & \textbf{35.0} \\
\hline
\multirow{6}{*}{4-shot} & MAE (Vit-B) \cite{He-CVPR-2022} & 45.1 & 83.3 & 57.1 & 89.8 & \textbf{47.4} & 87.3  & 26.8 & 62.5 & 20.1 & 58.0 \\
& CL-MAE (Vit-B) &  \textbf{45.5} &  \textbf{89.9} & \textbf{62.7} & \textbf{91.5} & \textbf{47.4} & \textbf{91.6} & \textbf{27.5} & \textbf{63.2}  & \textbf{21.8} & \textbf{58.8} \\
\cline{2-12}
& MAE (Vit-L) \cite{He-CVPR-2022} & 53.4 & 87.2 & 56.5 & 88.1 & 50.6 & 88.7  & 29.0 & 65.3 & 23.2 & 58.4 \\
& CL-MAE (Vit-L) &  \textbf{55.5} &  \textbf{91.8} & \textbf{61.8} & \textbf{90.4} & \textbf{52.7} & \textbf{89.9} & \textbf{31.9} & \textbf{66.4}  & \textbf{23.8} &  \textbf{59.4} \\
\cline{2-12}
& MAE (Vit-H)  \cite{He-CVPR-2022} & \textbf{53.7} & 90.8 & 58.1 & 89.9 & \textbf{51.4} & 89.3  & 31.0 & 65.7 & 23.0 & \textbf{58.5} \\
& CL-MAE (Vit-H) &  \textbf{53.7} &  \textbf{90.9} & \textbf{63.1} & \textbf{92.0} & 50.2 & \textbf{91.4}  & \textbf{31.8} & \textbf{67.0} & \textbf{23.3} & \textbf{58.5} \\
\hline
\multirow{6}{*}{8-shot} & MAE (Vit-B) \cite{He-CVPR-2022} & 57.6 & 95.9 & 65.9 & 97.8 & 56.5 & \textbf{96.6} & 35.7 & 70.6  & 29.7 & 65.5 \\
& CL-MAE (Vit-B) & \textbf{59.0} &  \textbf{96.5} & \textbf{69.3} & \textbf{98.2} & \textbf{57.3} & 96.3  & \textbf{36.0} & \textbf{71.5} & \textbf{32.4} & \textbf{69.4} \\
\cline{2-12}
& MAE (Vit-L) \cite{He-CVPR-2022} & 61.1 & 97.2 & 67.1 & 97.9 & \textbf{60.0} & 97.4  & 39.9 & 74.7 & 32.9 & 68.1 \\
& CL-MAE (Vit-L) & \textbf{63.1} & \textbf{97.4} & \textbf{70.3} & \textbf{98.7} & 59.8 & \textbf{97.7}  & \textbf{43.0}  & \textbf{76.3} & \textbf{33.6} & \textbf{69.4} \\
\cline{2-12}
& MAE (Vit-H)  \cite{He-CVPR-2022} & \textbf{62.4} & \textbf{96.3} & 68.3 & 97.9 & 62.1 & 95.7  & 41.9 & 76.2 & 35.8 & 70.1 \\
& CL-MAE (Vit-H) & \textbf{62.4} & 96.2 & \textbf{70.3} & \textbf{98.6} & \textbf{63.1} & \textbf{96.6}  & \textbf{43.3} & \textbf{76.5} & \textbf{37.8} & \textbf{72.1} \\
\hline
\multirow{6}{*}{16-shot} & MAE (Vit-B) \cite{He-CVPR-2022} & 53.3 & 93.7 & 68.1 & 97.4 & 52.7 & \textbf{93.1}  & 40.1 & 73.9 & 28.4 & 67.6 \\
& CL-MAE (Vit-B) & \textbf{57.9} & \textbf{96.8} & \textbf{72.3} & \textbf{99.0} & \textbf{54.2} & 92.7  & \textbf{41.0} & \textbf{76.4} & \textbf{31.1} & \textbf{69.0} \\
\cline{2-12}
& MAE (Vit-L) \cite{He-CVPR-2022} & 64.5 & 95.1 & 70.1 & 97.8 & 61.3 & 95.4  & 47.0 & 81.1 & 35.5 & 73.7 \\
& CL-MAE (Vit-L) & \textbf{66.0} & \textbf{97.3} & \textbf{73.3} & \textbf{99.1} & \textbf{62.1} & \textbf{95.5}  & \textbf{49.2} & \textbf{81.9} & \textbf{36.2} & \textbf{75.3} \\
\cline{2-12}
& MAE (Vit-H)  \cite{He-CVPR-2022} & 68.1 & 96.7 & 72.5 & 98.2 & 66.3 & \textbf{96.7}  & 51.3 & 82.9 & 42.6 & 76.9 \\
& CL-MAE (Vit-H) & \textbf{69.0} & \textbf{98.1} & \textbf{74.4} & \textbf{99.1} & \textbf{67.2} & 96.6  & \textbf{51.8} & \textbf{82.7} & \textbf{43.2} & \textbf{78.1} \\
\hline
\end{tabular}
\vspace{-0.2cm}
\caption{Few-shot linear probing results on five benchmarks: Aerial Images, Airbus Wind Turbines, Architectural Heritage Elements, Sea Animals, and Sport Balls. The results are reported for MAE \cite{He-CVPR-2022} and CL-MAE (ours) based on various backbones (ViT-B, ViT-L, ViT-H).  The reported accuracy rates are averaged over three runs. The top scores for each backbone on each data set are in bold.}
\label{table:fewshot_results} 
\end{table*}



\noindent
\textbf{Effect of diversity loss on masking.} 
Figure \ref{fig:diversity_loss} shows the results before and after introducing the diversity loss ($\mathcal{L}_{div}$) as an objective of our learnable masking module. When the masking module is trained without adding the diversity loss, the module enters mode collapse, producing nearly identical masks, regardless of the input sample. This can be problematic, as CL-MAE might overfit to the task and learn to reconstruct only a certain configuration of patches. In contrast, integrating the diversity loss helps our module to escape mode collapse and generate diverse masks, which solves the problem of overfitting to certain mask configurations.

\noindent
\textbf{Effect of Kullback-Leibler loss on masking.} 
Figure \ref{fig:kl_loss} illustrates three masks generated by our masking module during training, illustrating the effect of removing the Kullback-Leibler loss. At the beginning of the training process, when the module aims to make the reconstruction task easy for the MAE, it learns to leave all patches visible. As the training progresses, the module starts behaving like an adversary, aiming to make the reconstruction task hard for the MAE. From this point on, our module starts masking a number of tokens until it converges to masking all tokens. At this point, the MAE has literally no chance at reconstructing the input, being unable to further learn any useful information. To mitigate this issue, we add the Kullback-Leibler loss, which ensures the number of masked tokens complies with the desired ratio given as hyperparameter, irrespective of the complexity of the task. 

\noindent
\textbf{Few-shot linear probing results.} 
In Figure \ref{fig:fewshot_results}, we present few-shot linear probing results on Aerial Images \cite{Xia-TGRS-2017}, Airbus Wind Turbines \cite{Airbus-Wind}, Architectural Heritage Elements \cite{AHE_dataset}, Sea Animals \cite{SeaAnimal_dataset}, and Sport Balls \cite{SportsBalls_dataset} data sets. We complement the graphs illustrated in Figure \ref{fig:fewshot_results} with the results reported in Table \ref{table:fewshot_results}. These experiments are meant to quantify the transferability of the self-supervised representations learned by MAE \cite{He-CVPR-2022} and CL-MAE in the few-shot learning scenario. In 138 out of 150 cases, CL-MAE outperforms MAE, with absolute gains varying between $+0.1\%$ and $+13.5\%$. We generally observe that CL-MAE tends to bring higher gains for the 1-shot and 2-shot scenarios, especially when the ViT-L and ViT-H backbones are applied on the Aerial Images \cite{Xia-TGRS-2017}, Airbus Wind Turbines \cite{Airbus-Wind}, Architectural Heritage Elements \cite{AHE_dataset}, and Sea Animals \cite{SeaAnimal_dataset} data sets. Overall, the few-shot experiments confirm the observations on the nearest neighbor and linear probing experiments presented in the main article. We thus conclude that our curriculum learning approach represents a useful addition to the MAE framework.

\end{document}